\documentclass[letterpaper, 10 pt, conference]{ieeeconf}  % Comment this line out if you need a4paper

\IEEEoverridecommandlockouts                              % This command is only needed if 
\usepackage{graphicx} % for pdf, bitmapped graphics files
\usepackage{algorithm}
\usepackage{algorithmic}
\usepackage{float}
\usepackage{subfig}
\usepackage{url}
\usepackage[noadjust]{cite}

\usepackage{amsmath} % assumes amsmath package installed
\usepackage{amssymb}  % assumes amsmath package installed

\usepackage[dvipsnames]{xcolor}

\title{\LARGE \bf
Global Uncertainty-Aware Planning for Magnetic Anomaly-Based Navigation
}

\author{Aditya Penumarti$^{1}$ and Jane Shin$^{1}$% <-this % stops a space
\thanks{$^{1}$Aditya Penumarti and Jane Shin are with the Department of Mechanical and Aerospace Engineering,
        University of Florida, Gainesville, FL, USA
        {\tt\small apenumarti,jane.shin@ufl.edu}}%
}
\begin{document}

\maketitle
\thispagestyle{empty}
\pagestyle{empty}

\begin{abstract}
Navigating and localizing in partially observable, stochastic environments with magnetic anomalies presents significant challenges, especially when balancing the accuracy of state estimation and the stability of localization. Traditional approaches often struggle to maintain performance due to limited localization updates and dynamic conditions. This paper introduces a multi-objective global path planner for magnetic anomaly navigation (MagNav), which leverages entropy maps to assess spatial frequency variations in magnetic fields and identify high-information areas. The system generates paths toward these regions by employing a potential field planner, enhancing active localization. Hardware experiments demonstrate that the proposed method significantly improves localization stability and accuracy compared to existing active localization techniques. The results underscore the effectiveness of this method in reducing localization uncertainty and highlight its adaptability to various gradient-based navigation maps, including topographical and underwater depth-based environments.
\end{abstract}

\section{Introduction}

Accurate and reliable localization of autonomous vehicles remains a fundamental challenge in various field applications.
While the Global Positioning System (GPS) can provide precise positioning, GPS has become less reliable or unavailable in some applications, such as indoor navigation or long-duration operations in expansive aerial and underwater environments.
Relying on inertial measurement units (IMUs) introduces a challenge from cumulative drift errors over time. Visual or feature-based localization techniques, such as Simultaneous Localization and Mapping (SLAM), have proven effective for indoor navigation, however, their application in aerial and underwater long-duration missions is limited, largely due to the scarcity of distinct features in these large field environments and their sensitivity to environmental conditions. Thus, more robust and reliable localization solutions are needed in these domains.

\begin{figure}[!ht]
    \centering
    \includegraphics[width=\linewidth]{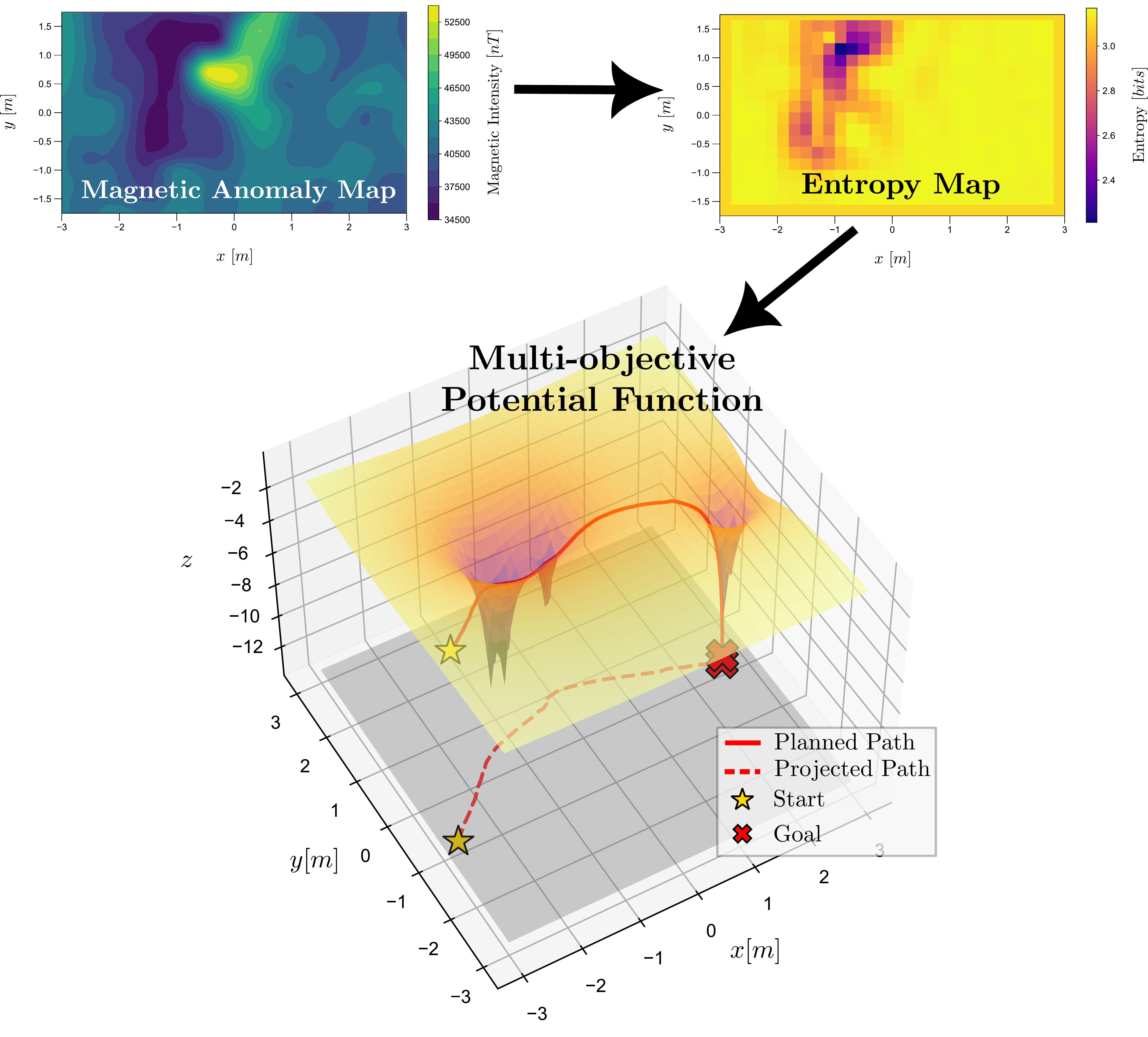}
    \caption{Pipeline for the Informative Potential Field Global Planner. The magnetic anomaly map, characterized by rich gradients, is used to compute an entropy map. This entropy map enables the application of a multi-objective potential function that balances both information gain and progress toward a specified goal. The resulting multi-objective path is depicted in the 3D plot, where the planned path is shown over the potential field, along with its projection onto a 2D plane as a dotted line.}
    \label{fig:representative_figure}
\end{figure}

Magnetic-based navigation (MagNav) presents a promising solution in these scenarios by leveraging the robustness and uniqueness of Earth's magnetic field, which is globally available and robust to external interference \cite{group_magnetic_2002}.
Magnetic-based localization leverages the spatial frequency features of the given magnetic field map to estimate position, as demonstrated in both ground-based \cite{storms_magnetic_2010, shockley_ground_2012, ramos_information-aware_2022}, aerial \cite{canciani_absolute_2016, canciani_airborne_2017, canciani_magnetic_2022}, and underwater \cite{quintas_auv_2019,jung_navigation_2020} vehicle applications across indoor and outdoor environments. This magnetic-based localization approach has also been adapted for use with SLAM to address scenarios where a pre-existing map is not available, as shown in \cite{vallivaara_simultaneous_2010} and \cite{lee_magslam_2020}, although the results suggest additional sensors may help for effective localization.

\begin{figure}[ht!]
    \centering
    \subfloat[\label{fig:max_info_diff_yaws}]{
        \includegraphics[width=\linewidth]{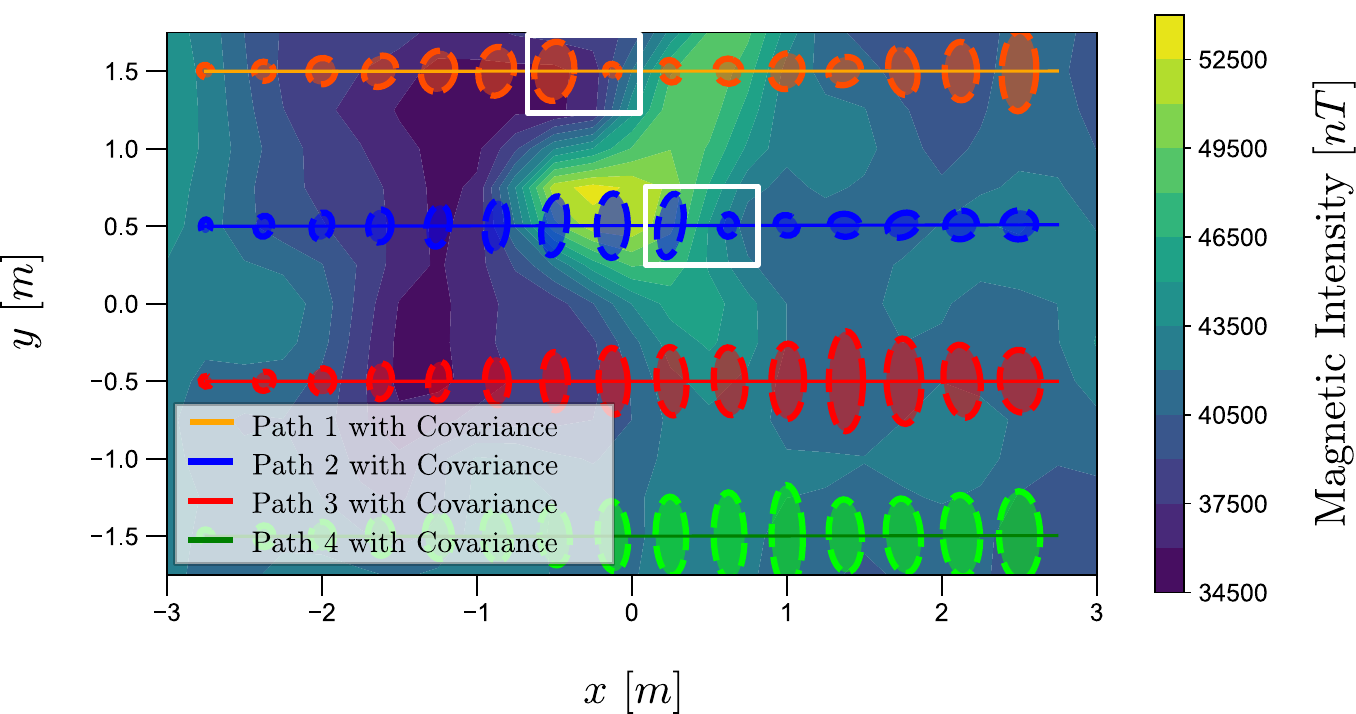}}
    \vfill
    \subfloat[\label{fig:ent_map_diff_yaws}]{
        \includegraphics[width=\linewidth]{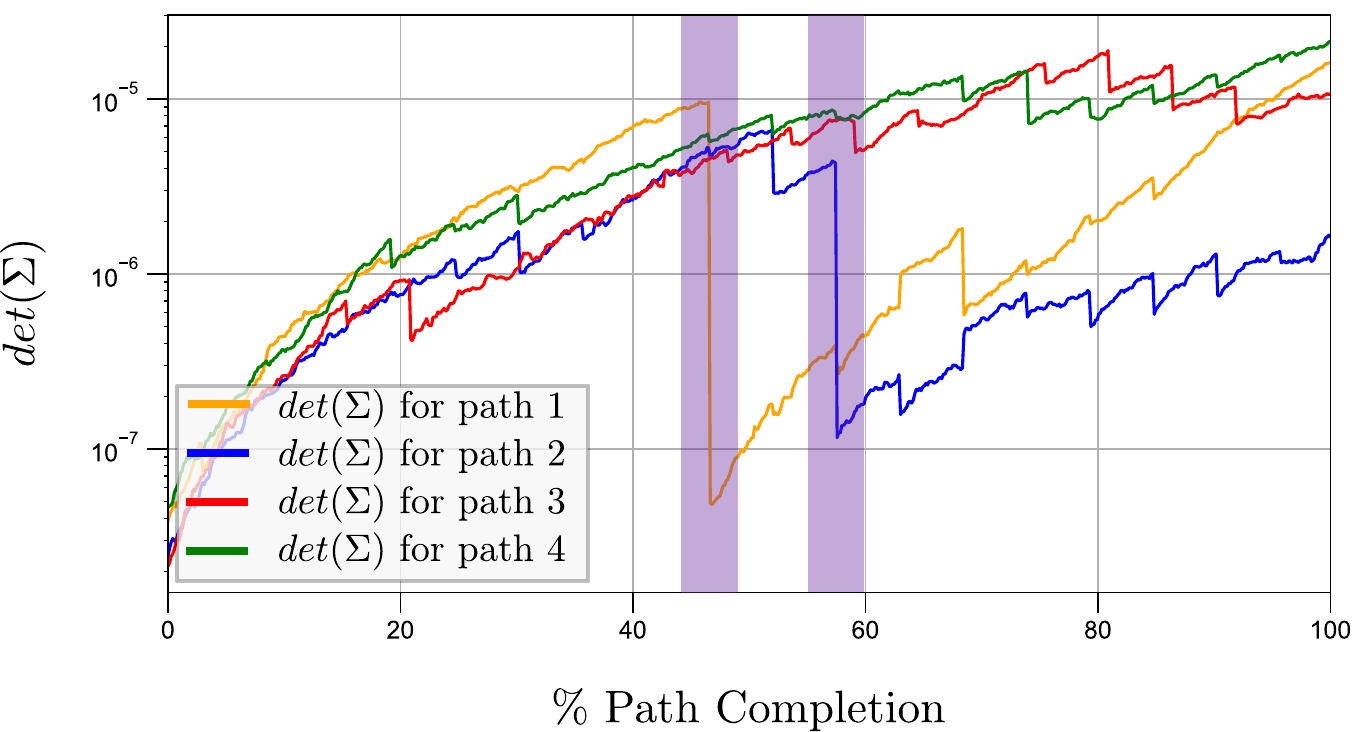}}
    \caption{(a) Simulated paths with localization uncertainty ellipses over a magnetic anomaly map. Paths through high-intensity variation reduce uncertainty, while low-variation paths show growing covariance. White regions correspond to areas where uncertainty is reduced. (b) Determinant of localization uncertainty for the paths in (a), with highlighted areas corresponding to the white regions of reduced uncertainty in (a).}
    \label{fig:preliminary}
\end{figure}

While MagNav has demonstrated effectiveness for GPS-denied navigation, it brings its unique challenges:
localization uncertainty is directly dependent on the spatial variability of the magnetic field (Fig. \ref{fig:preliminary}(b)).
Therefore, an effective and robust MagNav system should actively adapt the path to maintain low localization uncertainty while achieving its original mission objectives. There are a few studies that have explored this active localization for magnetic navigation. In \cite{kemppainen_magnetic_2015}, the SLAM exploration algorithm is developed using block partitioning and depth-first search to maximize information gain, but it focuses on local exploration, limiting global information distribution for long-term planning. Similarly, \cite{ramos_information-aware_2022, penumarti_real-time_2024} proposed active localization methods based on expected entropy reduction and the observability Gramian, enabling real-time path planning but still constrained in exploiting information-rich regions globally.

This paper presents a novel global path planning approach for active localization in magnetic-based navigation by formulating information gain based on the spatial frequency of the given magnetic field map and by constructing potential functions to enable multi-objective planning.
For representing spatial frequency for localization, we have developed an entropy map, inspired by \cite{vieux_dem_1993}, where entropy is used to evaluate aggregate information loss in smoothed terrain models.
For multi-objective planning, we have constructed a potential function to consider active localization for magnetic navigation in addition to the original missions.
As an example, we studied a scenario of performing localization uncertainty reduction while performing the original path following.
Hardware experiments have been performed to demonstrate the feasibility and effectiveness of the proposed global multi-objective planning algorithm.

\subsection{Contributions}
First, the presented method has formulated an information function, called entropy map, that can represent the amount of localization uncertainty reduction associated with the given magnetic field map's spatial features. Second, unlike traditional magnetic-based planning approaches that are often computationally intensive or myopic, the proposed method provides a global path optimization thanks to the properties of entropy map. Third, the design of a potential function to consider multi-objective planning enables simple and global expansion from existing planning schemes. Lastly, the contributions are rigorously validated through hardware experiments to confirm the practical significance and effectiveness of the proposed methods. This comprehensive approach not only demonstrates the practical applicability of the proposed methods but also contributes to the theoretical understanding of information in gradient-based maps for different domains.

\section{Preliminaries}
This section describes how magnetic-based localization is heavily dependent on the given environment via a magnetic anomaly map, highlighting the necessity of active localization. As shown in Fig. \ref{fig:preliminary}, a test is performed to compare magnetic-based localization performance over four distinct vehicle trajectories.
Trajectories passing through areas of high-intensity magnetic variation (orange and blue paths) showed a rapid collapse in localization covariance; specifically, the orange path passed through a dense intensity gradient, indicating improved localization performance in these regions. In contrast, trajectories through areas with low-intensity variation resulted in significantly degraded localization performance. These results indicate that high-intensity variation areas on the magnetic anomaly map provide lower entropy for the localization process, inspiring the formulation of an entropy map to represent localization uncertainty reduction in this paper.

\section{Problem Formulation}
This paper considers a global path planning problem for a mobile robot tasked with navigating in a GPS-denied environment while minimizing its localization uncertainty using a hybrid scalar-vector magnetometer. The workspace is defined by a 2-dimensional space $\mathcal{W}\subset \mathbb{R}^{2}$, and the magnetic field map over $\mathcal{W}$ is denoted by $m~:~\mathcal{W}\rightarrow \mathbb{R}^{+}$. The magnetic field map $m$ is assumed known \textit{a priori}, for example, from pre-surveying, and static. Localization is performed solely based on the onboard magnetometer measurement and a given magnetic field map, without other sensors like an internal measurement unit.

% robot state
The inertial frame $\mathcal{F}_{W}$ with $x_{I}y_{I}z_{I}$-axes is defined in $\mathcal{W}$, and the body-fixed frame $\mathcal{F}_{B}$ with $x_{B}y_{B}z_{B}$-axes is defined and fixed on the mobile robot. The robot position and orientation are defined by the position and orientation of $\mathcal{F}_{B}$ concerning $\mathcal{F}_{W}$. In this paper, we assume that the mobile robot is a differential-drive ground robot, although this assumption can be easily generalized by considering other dynamics constraints. Then, the position of the robot is defined by $x$ and $y$ position of $\mathcal{F}_{B}$ in $x_{I}y_{I}$-plane, and the orientation of the robot is defined by $\theta$, which is the angle of $x_{B}$-axis concerning $x_{I}$-axis. By denoting a certain time step by $k=0,1,2,\dots$, a state vector that represents the robot pose at time $k$ is defined by $\mathbf{x}_{k}=[x_{k}~y_{k}~\theta_{k}]^{T}$.

% robot motion model
By denoting the control state vector at time step $k$ as $\mathbf{u}_{k}$, the robot's motion model is denoted by
\begin{equation}
    \mathbf{x}_{k+1} = \mathbf{f}\left(\mathbf{x}_{k}, \mathbf{u}_{k}\right)
    \label{eq:motion_model}
\end{equation}
In this letter, the motion model of a differential-drive robot can be derived from a unicycle model. By letting $V$ and $\omega$ be the linear and angular velocities, the kinematics of a unicycle model gives
\begin{align}
    x_{k+1} &= x_{k} +V\cos \theta \Delta t + \epsilon_{x}\\
    y_{k+1} &= y_{k} + V \sin \theta \Delta t + \epsilon_{y}\\
    \theta_{k+1} &= \theta_{k} + \omega \Delta t + \epsilon_{\theta}
\end{align}
where $\Delta t$ is the size of each time step and $\epsilon_{x}$, $\epsilon_{y}$, $\epsilon_{\theta}$ are random Gaussian noises with zero mean and standard deviations $\sigma_{x}$, $\sigma_{y}$, and $\sigma_{\theta}$, respectively. The robot's control input is defined by $\mathbf{u}_k = \left[V~\omega \right]^{T}$. This motion model can be represented by a probability distribution $p(\mathbf{x}_{k+1}|\mathbf{x}_{k},\mathbf{u}_{k})$.

The sensor measurement model represents the process of generating magnetometer measurements. The magnetometer reading obtained from the state $\mathbf{x}_{k}$ is denoted by $z_{k}\in\mathbb{R}^{+}$. By assuming that this reading is only dependent on the earth's magnetic field and the robot's current pose, the sensor model can be represented by
\begin{equation}
    z_{k} = g(\mathbf{x}_{k},m)
    \label{eq:sensor_model}
\end{equation} 
The measurement model can be defined by a probability distribution conditioned on the target state $\mathbf{x}_{k}$ and the map $m$ and denoted by $p(z_{k}|\mathbf{x}_{k},m)$. As an example, one can model the sensor measurement model as a Gaussian by $z_{k} \sim \mathcal{N}\left(m(\mathbf{x}_{k}),\sigma^2\right)$, which represents a magnetometer that obtains the magnetic field intensity at $\mathbf{x}_{k}$ with Gaussian noise with variance $\sigma^2$.
The robot's pose was determined at each time step using a particle filter following \cite{thrun_probabilistic_2005}. A particle filter is ideal for use in MagNav since it lends itself to use in a temporal varying sensor measurement and the ability for a magnetic map to be interpolated \cite{canciani_absolute_2016}. 

A global planner must account for information across the entire map, but evaluating localization entropy over such a large area is computationally expensive, even in a discretized space. An alternative approach is necessary to efficiently incorporate entropy into the planning process. Drawing inspiration from methods used to assess digital elevation model (DEM) quality, this approach allows the planner to identify key points that minimize the robot's localization uncertainty by relying solely on the magnetic map. As a result, the robot can navigate toward areas of low entropy that a local motion planner might otherwise overlook, improving overall localization performance.

\section{Global Planner for Entropy Maps}
\subsection{Entropy Maps} \label{sec:entropy_map}
The entropy for a discrete space is defined as:
\begin{equation}\label{eq:entropy_discrete}
    H(x)=\sum-p(x)\log(p(x)).
\end{equation}
where $p(x)$ represents the probability of a given random variable $x$.
To generate an "Entropy Map," a magnetic map must first be collected and interpolated, if necessary. This interpolation step is crucial because it enhances the level of detail in the magnetic field gradients, thereby improving the planner’s overall performance. Once the interpolation is complete, the magnetic map is segmented into bins to calculate entropy. As a preliminary step, the magnetic field measurements are normalized across the full range of intensities to ensure consistency in the entropy calculation \cite{chen_local_2018}, \cite{li_quantitative_2002}:
\begin{equation}\label{eq:bin_normalization}
    z_b^{[i]} = \frac{z_b^{[i]}-\min z_b}{\max z_b - \min z_b},
\end{equation}
where $z_b$ represents the binned magnetic intensities. To calculate entropy, a sliding window approach is employed to calculate entropy over a window of bins \cite{wise_information_2012}:
\begin{equation}
    W = \frac{m(a,b)}{\sum m(a,b)},
\end{equation}
here, $W$ represents the matrix of normalized values within the window, with $m(a,b)$ where $(a,b) \in (i,i+r) \times (j,j+r);\,i,j,r \subseteq \mathbb{Z}$, and $r$ is the window size. Typically, these values are summed over the entire map to compute the map's entropy. However, the goal here is to focus on calculating the local intensity curvature entropy. These normalized values are treated as statistical probabilities to compute the entropy using \eqref{eq:entropy_discrete}:
\begin{equation}
    H_W = \sum_{i=1}^N -p_i \log(p_i),
\end{equation}
where $H_W$ is the entropy of the window, and $p_i$ is the $i$-the probability value in the window.

\subsection{Potential Field Planner}\label{sec:potential_field_planner}

Since the low-entropy points tend to cluster due to the gradient-based nature of the magnetic anomaly maps, a planner that can leverage this characteristic is desired. Therefore, a potential field planner is implemented to plan the path through low-entropy areas.

The potential field planner traditionally uses both attractive potentials (minima) and repulsive potentials (maxima) to plan paths \cite{hwang_potential_1992}. Repulsive potentials typically represent obstacles, while attractive potentials guide the robot to the goal. Here, only the attractive potential is used since no obstacles are detected during navigation. However, future work could extend this research to include obstacle avoidance using predefined maps.

The attractive potential for the desired goal is defined as:
\begin{equation}\label{eq:potential_field_att_goal}
    U_{att,G} = \frac{1}{2} \xi_G \rho_G (q)
\end{equation}
$\xi_G$ is a scaling factor, and $\rho_G$ is the norm distance from the current point to the goal. The scaling factor $\xi_G$ is given by:
\begin{equation}\label{eq:goal_weight}
    \xi_G = 10*e^{\frac{1}{\rho_G(q)}}
\end{equation}
The attractive potential for each entropy point is defined as:
\begin{equation}\label{eq:potential_field_att_entropy}
    U_{att,H} = \sum_{i=1}^{N} \frac{1}{2} \xi_H^{[i]} \rho_H^{[i]}
\end{equation}
where $N$ is the number of entropy points within five standard deviations from the mean, and $\rho_H^{[i]}$ is the distance to the $i$-th entropy point. The scaling factor $\xi_H^{[i]}$ is calculated as:
\begin{equation}\label{eq:entropy_weight}
    \xi_H^{[i]} = 0.5^{\frac{1}{\log_2( H_W^{[i]})}}
\end{equation}

\section{Experiments}
\begin{figure}[t]
    \centering
    \includegraphics[width=0.75\linewidth]{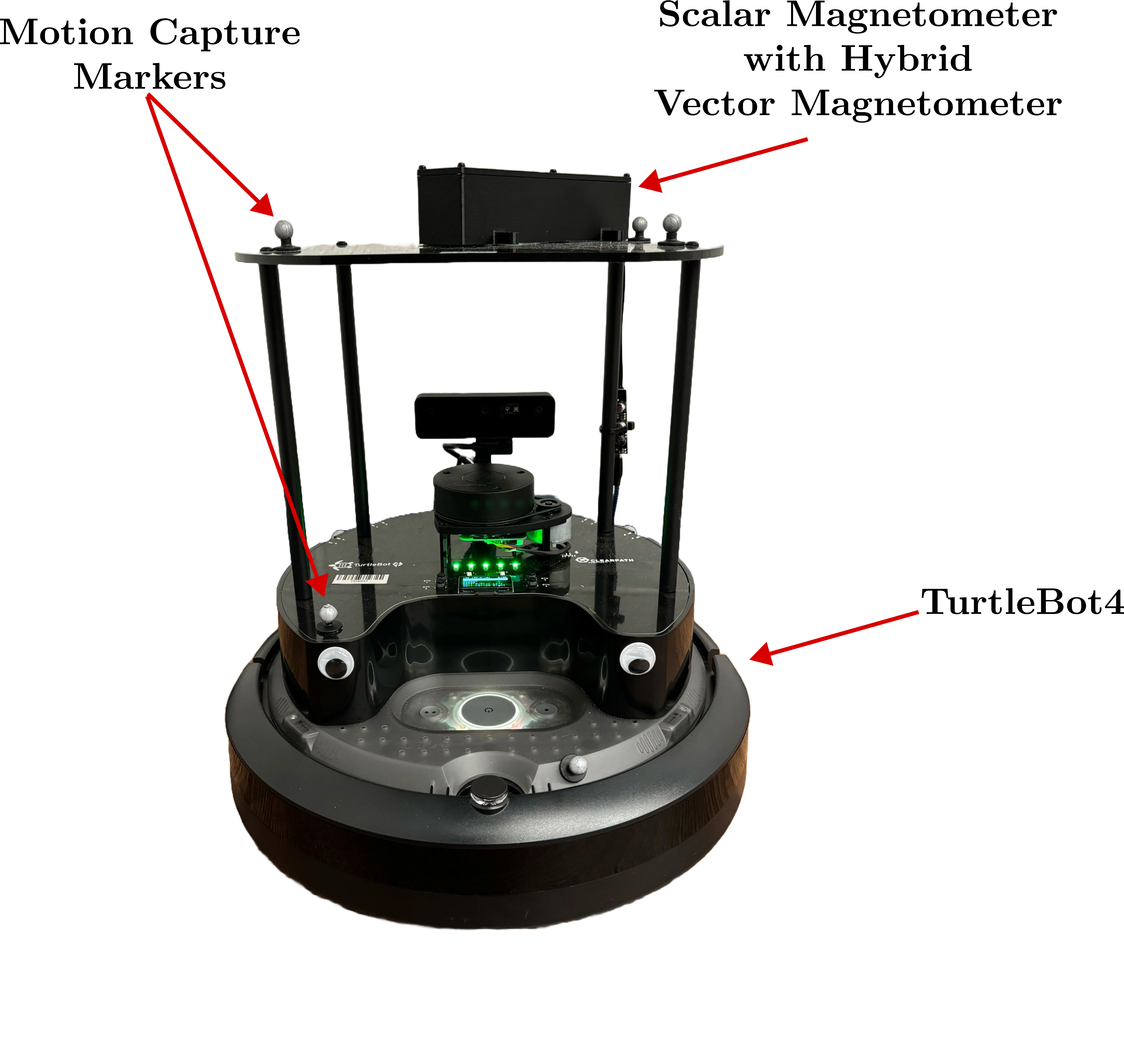}
    \caption{TurtleBot4 with a total field magnetometer with vector add-on and motion capture markers.}
    \label{fig:mag_setup}
\end{figure}
The entropy map path planning method was validated using a magnetic anomaly map collected at the University of Florida. A ClearPath Robotics TurtleBot4 was outfitted with a QUSpin QTFM total-field magnetometer, including a vector add-on, as shown in Fig. \ref{fig:mag_setup}. The map was collected on a $0.25\,\mathrm{m} \times 0.25\,\mathrm{m}$ spaced grid over the entire available workspace and correlated to a ground truth pose provided by a motion capture system. For the particle filter, that is used to estimate the state of the robot, the process covariance was set as the covariance matrix given by $Q =\boldsymbol{\mathrm{diag}}\left(x=(0.01\,\mathrm{m})^2,y=(0.01\,\mathrm{m})^2,\theta = (0.01745\,\mathrm{deg})^2\right)$. While the sensor covariance, $R$, was set as $(100\,\mathrm{nT})^2$

The motion planner by Penumarti et al. \cite{penumarti_real-time_2024} was evaluated against the method presented here under two comparative conditions: one considering high information gain and the other with no information considered in the receding horizon. The initial conditions were consistent across all methods, with tests conducted at two different yaw angles, $60^{\circ}$ and $90^{\circ}$.

\subsection{Entropy Map}
\begin{figure}[t]
    \centering
    \includegraphics[width=\linewidth]{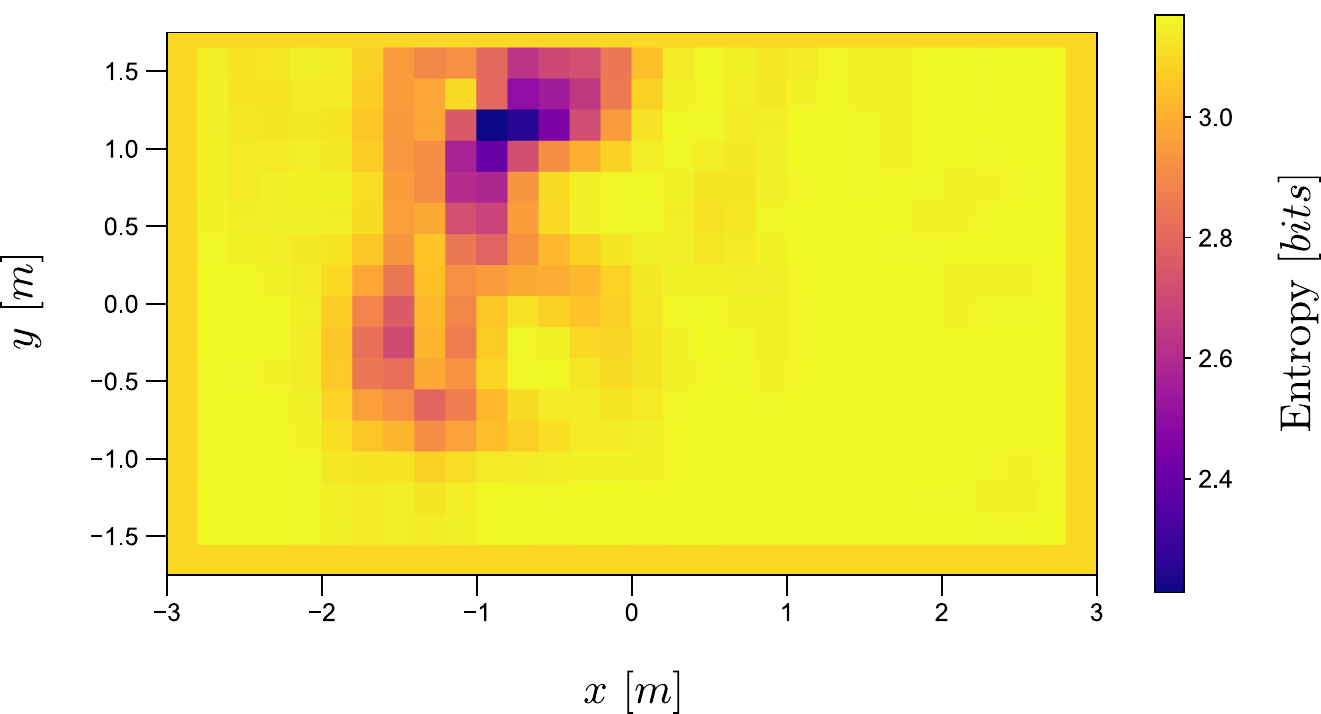}
    \caption{Entropy map generated from the magnetic anomaly map.}
    \label{fig:entropy_map}
\end{figure}

Following the methods outlined in Sec. \ref{sec:entropy_map}, an entropy map was generated with a bin size of $0.2\,\mathrm{m}$ and a window size of 2. The resulting map is shown in Fig. \ref{fig:entropy_map}. The purple areas in the map represent regions where the robot should navigate to gain the most information.

\subsection{Planning and Following}
\begin{figure}[t]
    \centering
    \subfloat[\label{fig:entropy_map_path_plan}]{
        \includegraphics[width=\linewidth]{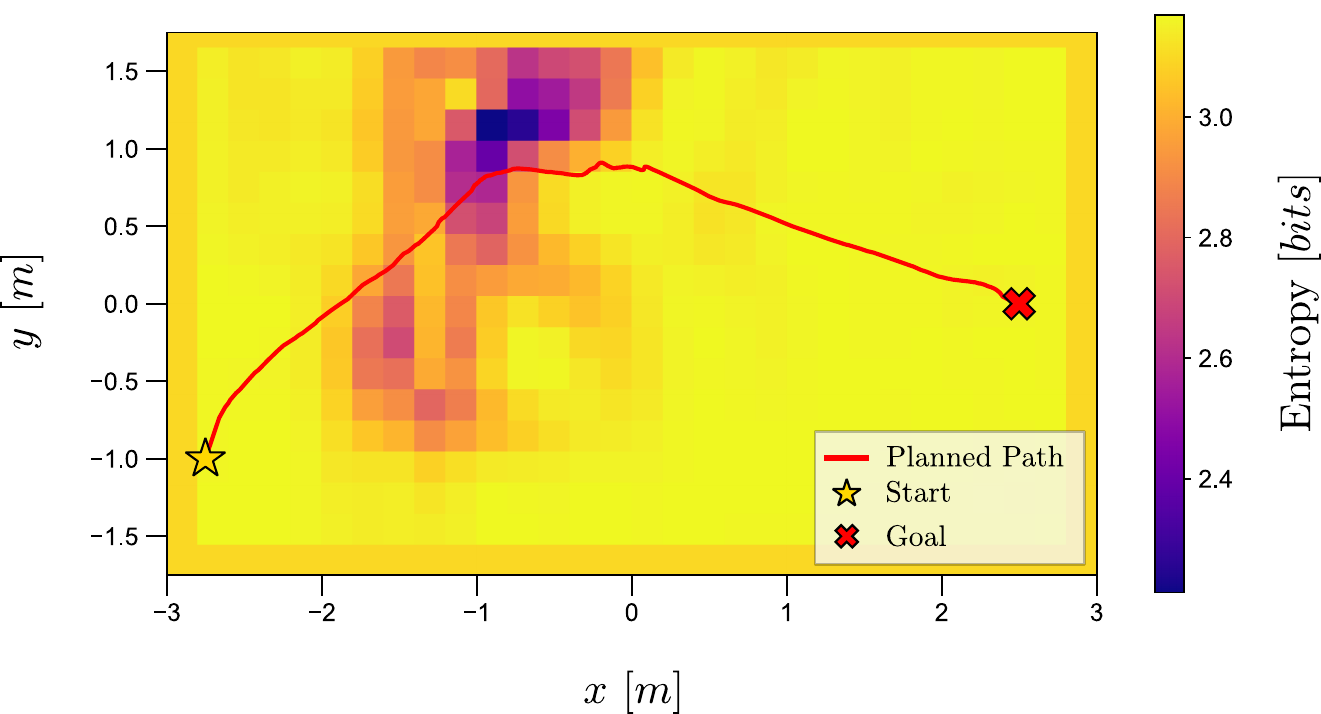}}
    \vfill
    \subfloat[\label{fig:mag_anomaly_path_plan}]{
        \includegraphics[width=\linewidth]{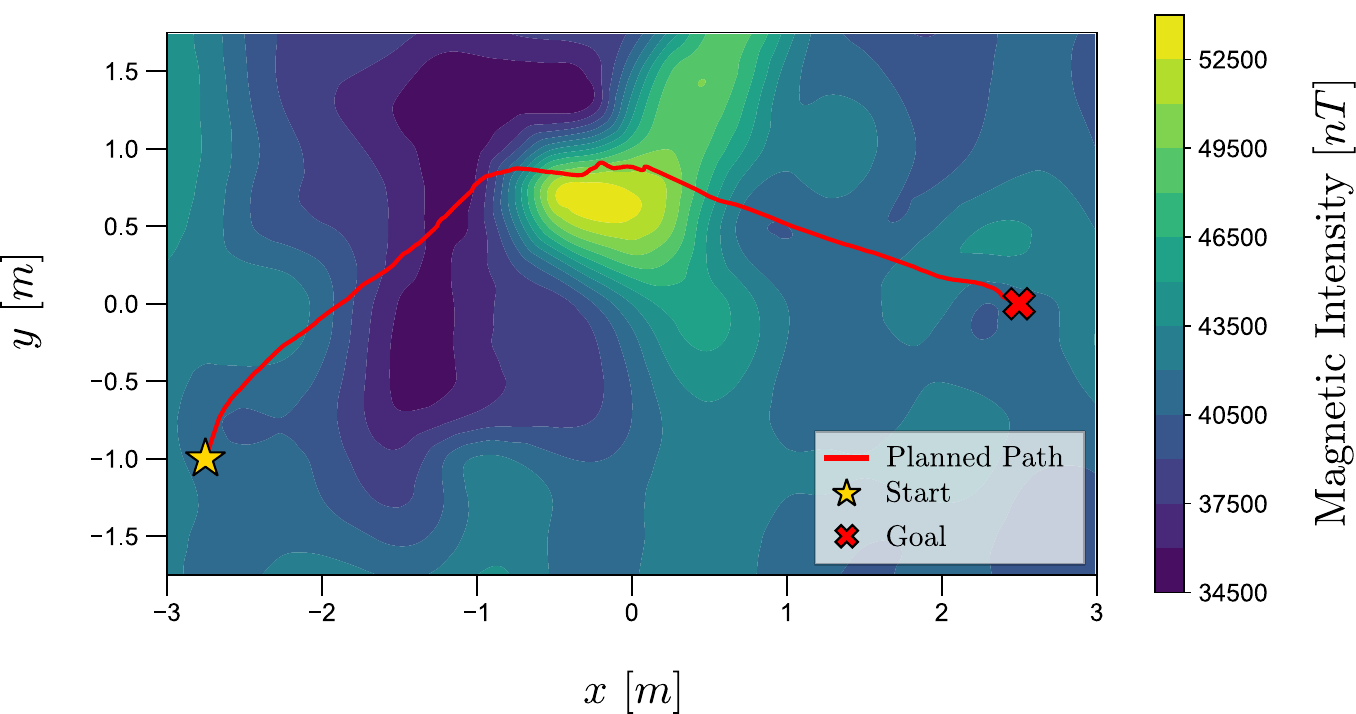}}
    \caption{(a) The planned path through the entropy map. (b) The planned path through the magnetic anomaly intensity map.}
    \label{fig:path_plan}
\end{figure}

Using the generated entropy map, a path was planned through the magnetic anomaly map. The start position was set at $[-2.75, -1.25],\mathrm{m}$ and the goal position at $[2.5, 0.0],\mathrm{m}$ to encourage deviations that take information into account. Given the dense information present in the map, only points with entropy values $5\sigma$ below the mean were selected, ensuring the selection of low-entropy points. While this method can be applied to the entire set of entropy points, the planner may become stuck in local minima depending on the structure of the prior map. To ensure a relatively smooth path, a moving average filter was applied. The generated paths are shown in Fig. \ref{fig:path_plan}. As observed in Fig. \ref{fig:path_plan}(b), the path follows large gradients in the magnetic map, favoring areas with significant gradients.

To follow a path, a simple Stanley controller \cite{hoffmann_autonomous_2007} was implemented. The Stanley controller generates angular twist (yaw rate) commands $\omega$ for a differential drive robot based on:
\begin{equation}\label{eq:stanley_controller}
    \omega = \theta_e + \arctan{\frac{k_s \delta_c}{v}}
\end{equation}
where $\theta_e$ is the angle error, found from the difference between the robot's estimated yaw and the angle to the closest point on the path, $\delta_c$ is the cross-track error found from the robot's estimated position and the closest point on the path, $v$ is the linear velocity, and $k_s$ is the Stanley gain.

The Stanley controller described in \eqref{eq:stanley_controller} was implemented and used to follow the planned path. A Stanley gain of $0.001$ was found to be appropriate for traversing the path, with a linear speed of $0.15\,\mathrm{m/s}$.
\subsection{Results}
\begin{figure}[t]
    \centering
    \subfloat[\label{fig:max_info_diff_yaws}]{
        \includegraphics[width=0.95\linewidth]{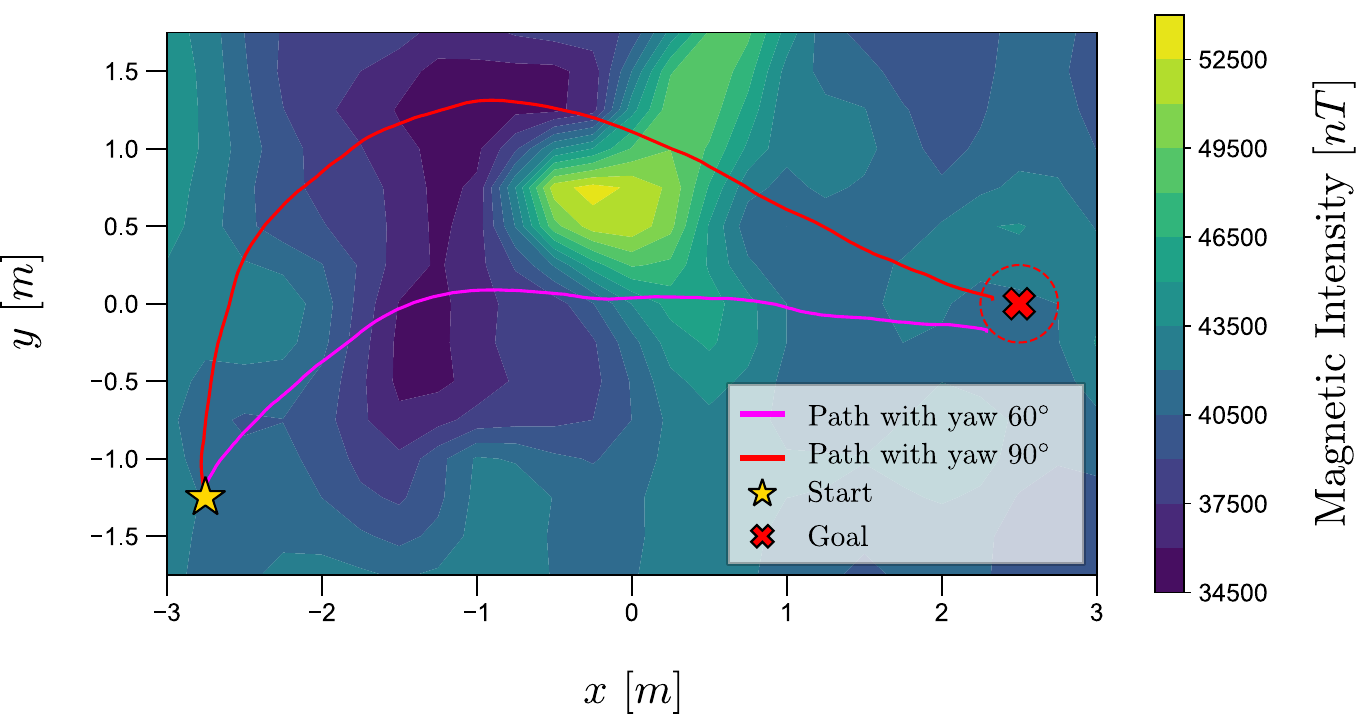}}
    \vfill
    \subfloat[\label{fig:ent_map_diff_yaws}]{
        \includegraphics[width=0.95\linewidth]{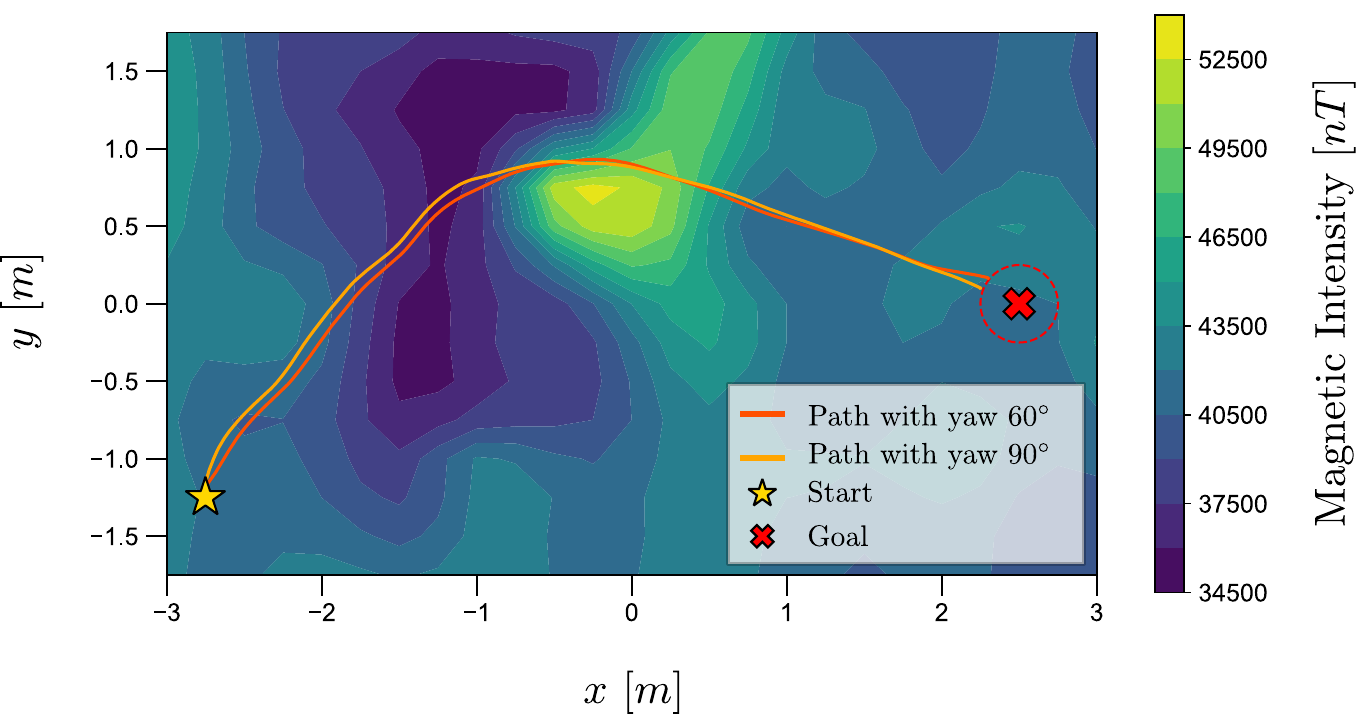}}
    \caption{(a) Paths of the information guided motion planner presented by Penumarti et. al with two different yaw angles, $60^{\circ}$ and $90^{\circ}$. (b) Paths of the entropy map potential field planner with the two different yaw angles.}
    \label{fig:diff_yaws}
\end{figure}

To demonstrate the reliance on initial conditions, the motion planner presented by Penumarti et al. was tested with two different yaw angles at the starting point. With a yaw of $60^{\circ}$, the navigation method tended to move through gradients in the middle of the map, whereas with a yaw of $90^{\circ}$, the robot navigated towards the upper gradients, as shown in Fig. \ref{fig:diff_yaws}(a). Similarly, the entropy map potential field planner was tested with the same yaw angles. The entropy map planner was robust to changes in initial conditions, following the same path in both instances, as shown in Fig. \ref{fig:diff_yaws}(b).

\begin{figure}[t]
    \centering
    \includegraphics[width=0.95\linewidth]{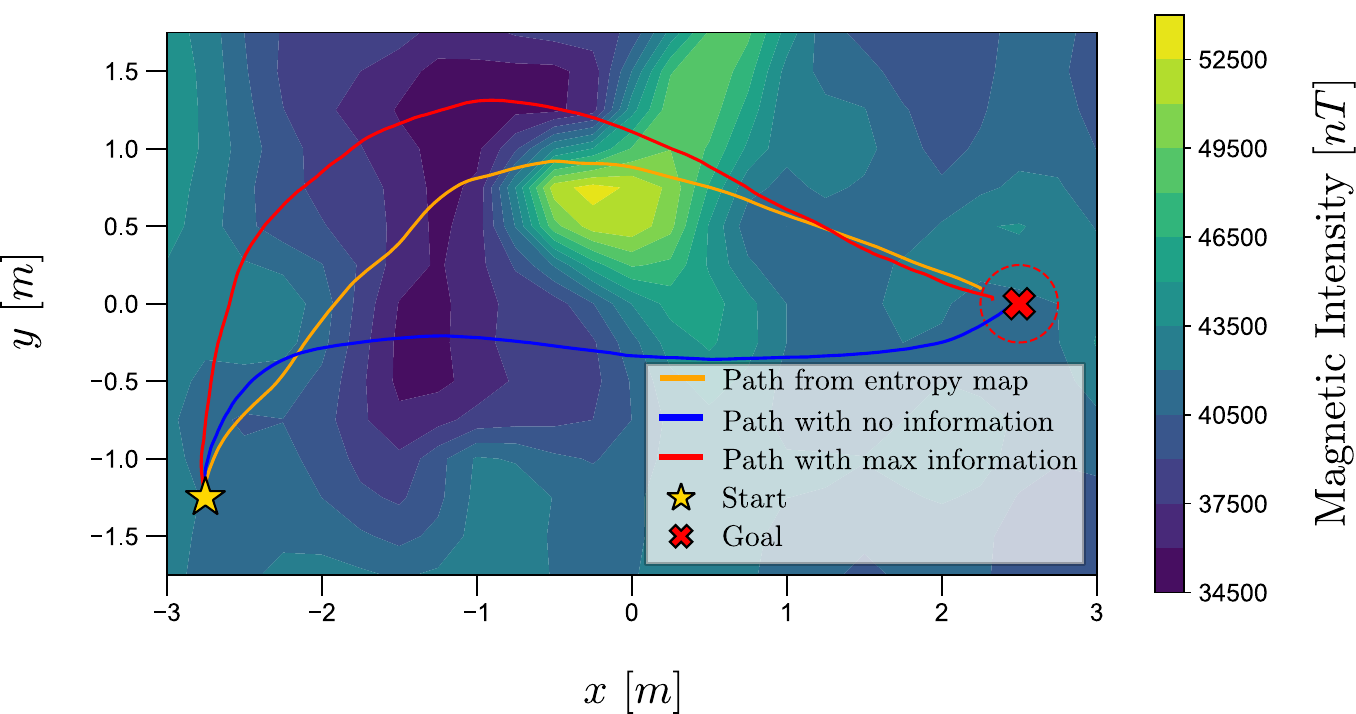}
    \caption{Robot paths to the goal using the motion planner with no information considered, high information considered, and the entropy map path planner.}
    \label{fig:robot_paths}
\end{figure}

The information-driven motion planner was tested under two conditions: one with high information gain and another without considering any information gain. These results were compared to the entropy map-based path planner. When no information gain was considered, the robot followed a direct path to the goal. When high information gain was considered, the robot navigated into the upper gradients but bypassed intermediate gradients. In contrast, the entropy map planner guided the robot through multiple high-information areas, maximizing information gain along the path. These robot paths are shown in Fig. \ref{fig:robot_paths}.

\begin{figure}[t]
    \centering
    \subfloat[\label{fig:max_info_diff_yaws}]{
        \includegraphics[width=0.9\linewidth]{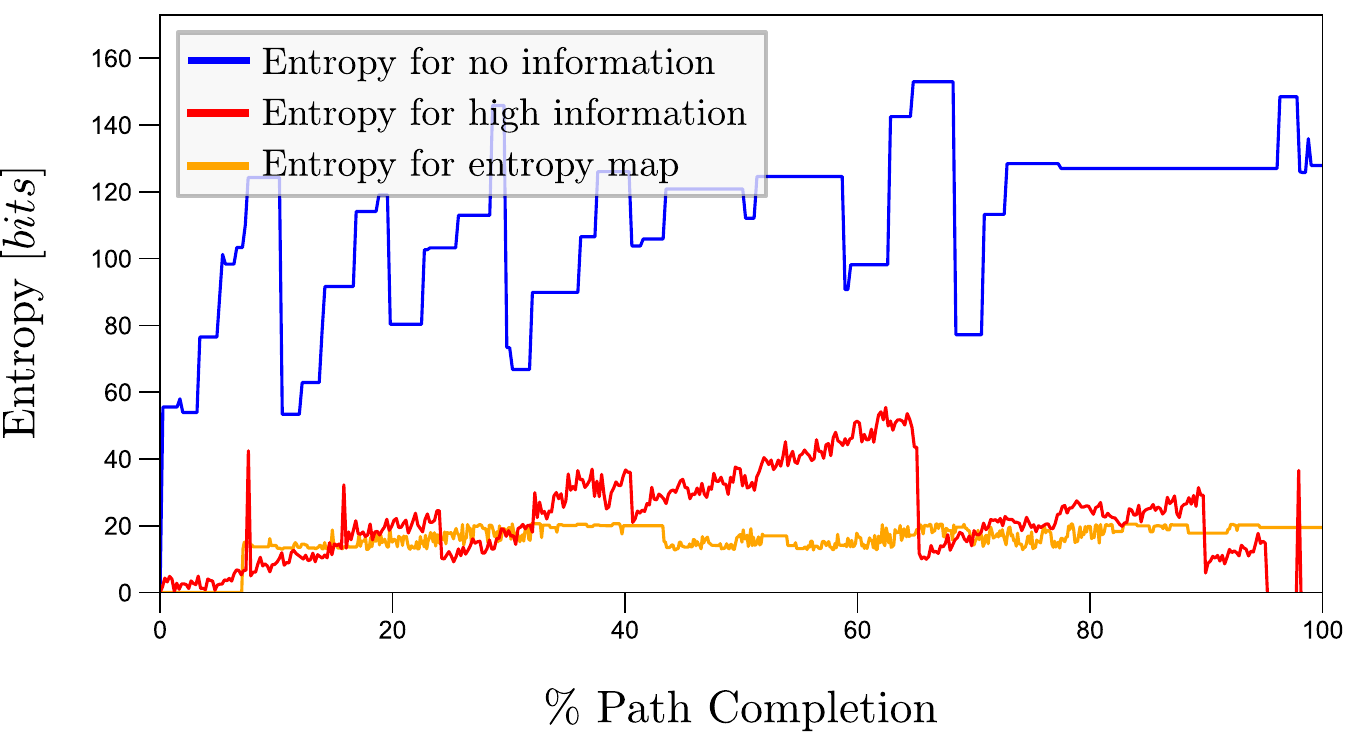}}
    \vfill
    \subfloat[\label{fig:ent_map_diff_yaws}]{
        \includegraphics[width=0.9\linewidth]{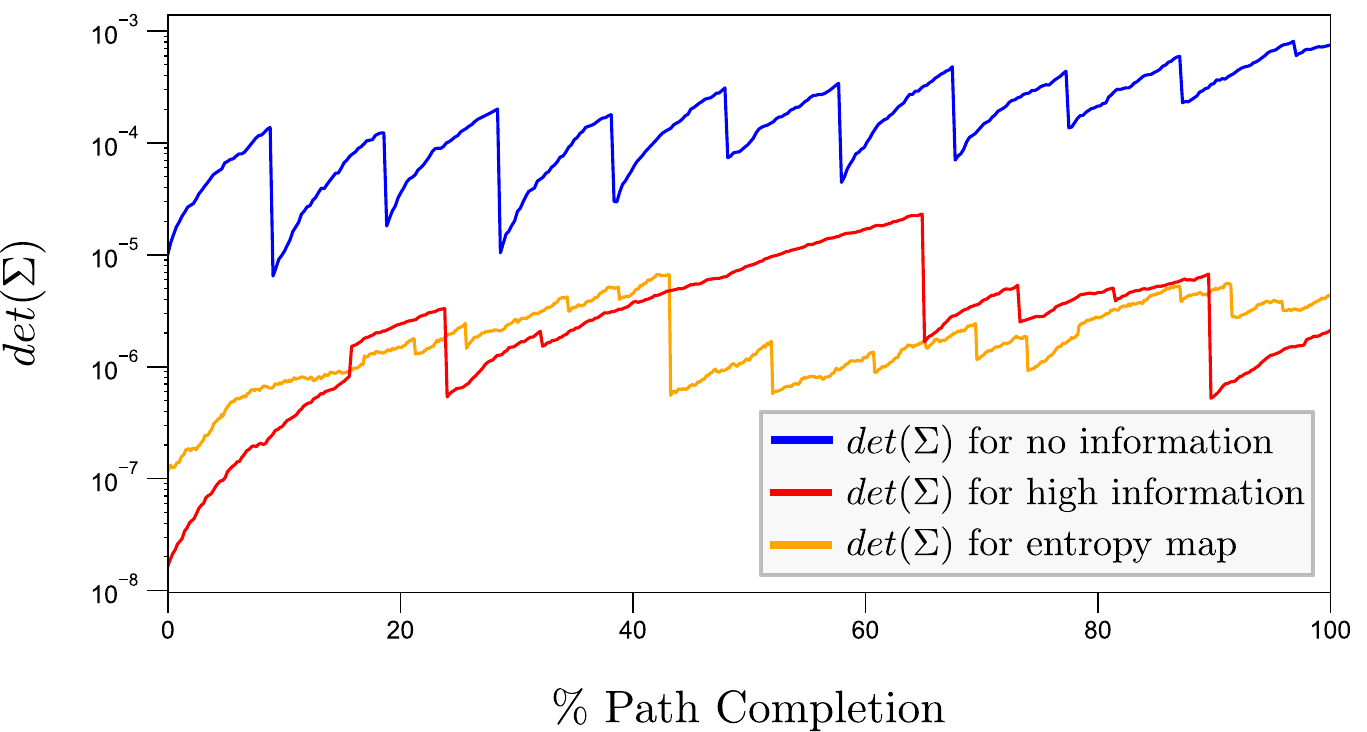}}
    \caption{(a) Entropy values along the robot paths shown in Fig. \ref{fig:robot_paths}. (b) Determinant of the covariance matrix along the same paths shown in Fig. \ref{fig:robot_paths}.
}
    \label{fig:entropy_covariance}
\end{figure}

Entropy values along the paths are depicted in Fig. \ref{fig:entropy_covariance}(a). The motion planner without information gain resulted in significantly higher entropy. The planner accounting for high information gain produced lower entropy, which decreased consistently. The entropy map planner maintained stable and lower entropy compared to the other two paths.

Considering the determinant of the covariance matrix for $x$, $y$, and $\theta$ (Fig. \ref{fig:entropy_covariance}(b)), the motion planner without information had a significantly higher and unstable covariance. The motion planner with information gain had lower covariance, but it fluctuated throughout the path. In contrast, the entropy map planner exhibited consistently lower and more stable covariance than the other methods.


\section{Discussion}
When comparing the entropy map to the magnetic anomaly map, it is evident that low-entropy points align with areas of high gradient. This is expected, as gradients represent regions of rapid change in magnetic intensity, which carry more information due to low-probability magnetic points \cite{canciani_absolute_2016}. The entropy map provides a statistical representation of spatial variability in magnetic intensity, akin to topographical maps. Through observation and experimentation \cite{canciani_airborne_2017,ramos_information-aware_2022,canciani_magnetic_2022,penumarti_real-time_2024}, a strong correlation is established between magnetic intensity gradients and information content.

The path planner, generated using the methods in Sec. \ref{sec:potential_field_planner}, initially deviates toward high-information points in the gradients, consistent with the weighting function in \eqref{eq:entropy_weight}. It also deviates toward some magnetic sinks on the left side of the map to maximize information gain, before being attracted to the goal by the secondary weighting function.

Observing the results with two different yaw angles, the entropy map planner was largely unaffected by changes in initial conditions, exhibiting only minor deviations in the early stages of the path. In contrast, the paths generated by the information-driven motion planner varied significantly depending on the initial yaw. This suggests that local-based planners, while robust in local information gain, may miss areas crucial to improving localization.

Comparing the three paths generated by the motion planner and entropy map planner, the differences are evident. The motion planner without information gain prioritized a direct route to the goal, disregarding localization enhancement. The planner with high information gain deviated into the upper map gradients but neglected others due to its local nature. Finally, the entropy map planner effectively navigated through multiple high-gradient areas, optimizing localization entropy. As shown in Fig. \ref{fig:entropy_covariance}(a), the entropy map planner maintains low entropy, suggesting more efficient navigation. Additionally, the covariance determinant, shown in Fig. \ref{fig:entropy_covariance}(b), confirms that the entropy map planner provides greater stability compared to the other planners.

\section{Conclusion}
This paper introduced a global multi-objective path planner for magnetic anomaly navigation (MagNav), demonstrating its effectiveness compared to a previously introduced information-aware navigation method. By utilizing entropy maps to assess spatial frequency variations in magnetic intensities across local map areas, the proposed global planner identified high-information regions that correlated with improved active localization performance. This approach not only enhanced the stability of localization uncertainty but also provided a versatile framework adaptable to other gradient-based navigation systems, including topographical and underwater depth-based scenarios.

Despite its advantages, the global planner is limited by its reliance on prior information and a preplanned path. Deviations such as noise or interference can impact its effectiveness. Therefore, this planner is best utilized as a complement to existing methods \cite{kemppainen_magnetic_2015,ramos_information-aware_2022,penumarti_real-time_2024}, where its integration can significantly reduce localization uncertainty. Future research should focus on developing a rigorous mathematical treatment to describe the relationship between spatial frequency in gradients and information content. Expanding the entropy map framework to other sensing modalities and incorporating multi-objective optimization could further enhance the adaptability and performance of the MagNav system.

\bibliographystyle{IEEEtran}
\bibliography{IEEEabrv, references}

\end{document}